\documentclass[12pt]{article}
\usepackage[T1]{fontenc}
\usepackage{amsmath,amssymb,amsthm}
\usepackage{graphicx}
\usepackage[margin=1in]{geometry}
\usepackage{booktabs}
\usepackage[authoryear,round]{natbib}
\usepackage[hidelinks]{hyperref}

\newcommand{\E}{\mathbb{E}}
\newcommand{\KL}{\mathrm{KL}}
\newcommand{\tr}{\operatorname{tr}}
\newcommand{\Cov}{\operatorname{Cov}}
\newcommand{\Var}{\operatorname{Var}}
\newcommand{\N}{\mathcal{N}}
\newcommand{\F}{\mathcal{F}}
\newcommand{\half}{\tfrac12}

\DeclareMathOperator{\diag}{diag}

\title{Stochastic Deep Learning:\\ \large A Probabilistic Framework for Modeling Uncertainty in Structured Temporal Data}
\author{James K Rice\thanks{PhD Candidate, University of Essex; james.k.rice@essex.ac.uk}}
\date{}

\begin{document}
\maketitle

\begin{abstract}
Stochastic Latent Differential Inference (SLDI) embeds an It\^o stochastic differential equation in the latent space of a variational autoencoder. Neural drift and diffusion coefficients describe how uncertainty evolves between irregularly spaced observations. I characterize the exact evidence gap that remains when the recognition model updates only the initial state, then use an observation-conditioned drift to allow subsequent measurements to inform the path. I also develop the original proposal to learn an auxiliary adjoint field. The field predicts coefficients of centered innovation terms and corrects the exact finite-grid gradient while preserving its expectation. A linear Gaussian example gives exact cancellation through second order. In nonlinear smoothing experiments I compare pilot budgets, vary the integration grid with coupled Brownian increments, and isolate the contributions of steps adjacent to observations. The results identify when tangent features improve the learned correction and when a lower-order fit is more effective.
\end{abstract}


\section*{Introduction}

Uncertainty is an intrinsic characteristic of real-world systems, from financial markets and climate dynamics to neural activity and disease progression. Traditional machine learning approaches, though highly flexible and performant, often provide point estimates with little or no quantification of uncertainty. This limitation can lead to overconfidence in predictions and reduced reliability in critical applications. To address this, probabilistic machine learning methods have emerged as a compelling alternative, aiming to integrate principled statistical reasoning with the representational power of deep learning.

Recent advances have seen the synthesis of probabilistic and deep learning paradigms, including Bayesian neural networks \citep{blundell2015weight}, Gaussian processes \citep{rasmussen2005gaussian}, and variational inference frameworks \citep{kingma2013auto}. However, a persistent challenge remains in modeling data whose uncertainty evolves with the latent state and between observation times. Standard methods may either ignore the dynamics of noise (by assuming static distributions) or struggle with scalability when integrated into deep architectures.

In this work, I propose Stochastic Latent Differential Inference (SLDI), a probabilistic framework that explicitly incorporates time-evolving uncertainty into deep generative modeling. My approach is grounded in It\^o stochastic differential equations (SDEs), which describe how a latent process evolves under random forcing. The encoder summarizes the observed sequence, the SDE propagates the latent state, and the decoder specifies how observations arise at the recorded times. This is the encoder--SDE--decoder framework of the first version \citep{rice2026sldi}. The present revision develops its two central components, the approximation to the posterior path distribution and the learned auxiliary field for backpropagation.

The restriction in the original encoder is easiest to see when the initial state is observed exactly. A later noisy observation of a Brownian trajectory still contains information about its intervening increments. Resampling the initial state cannot use that information, since the recognition model has already committed to the prior law of the increments. I quantify the attainable evidence lower bound (ELBO) under this restriction. The result helps distinguish an approximation error inherent to the chosen path family from an amortization error that could be reduced by improving the encoder. An observation-conditioned drift gives the same framework a way to respond to measurements throughout the path.

The auxiliary adjoint poses a different problem. The backward sensitivity on a sampled trajectory depends on later innovations. A learned approximation can therefore alter the expected update if it is inserted directly into the gradient or evaluated after the innovation it is supposed to correct. I fit the auxiliary field using exact sample sensitivities and evaluate it using information available before each innovation. The result is an additive control variate for the gradient of the same ELBO. The experiments use this distinction to measure variance at fixed parameters. A known linear Gaussian model verifies the exact cancellation, and nonlinear experiments ask how the correction is distributed in time when observations arrive irregularly.

\section*{Background and Literature Review}

Probabilistic approaches to machine learning aim to capture epistemic and aleatoric uncertainty in predictive modeling. Unlike deterministic neural networks, probabilistic models assign a distribution over parameters or latent variables, enabling a richer understanding of model confidence.

A foundational model in this space is the Bayesian neural network (BNN), in which posterior distributions over weights are learned, typically using approximate inference techniques such as variational inference or Markov Chain Monte Carlo \citep{neal2012bayesian}. While theoretically appealing, BNNs often suffer from computational overhead and convergence instability, motivating the use of alternatives like dropout-based approximations \citep{gal2016dropout} or deep ensembles \citep{lakshminarayanan2017simple}.

Variational autoencoders (VAEs) represent another major branch of probabilistic deep learning. VAEs use an encoder-decoder architecture to learn a low-dimensional latent space, where uncertainty is captured via a variational posterior \citep{kingma2013auto}. However, standard VAEs often model time using discrete latent states or ignore temporal coherence, which limits their utility in dynamic environments.

These distinctions matter when the latent dynamics are stochastic. Conditional on fitted network weights, SLDI represents uncertainty in the process path and in the decoder's observation model. A Bayesian treatment of the network weights would represent another source of predictive uncertainty. Both can matter, for instance when measurement noise is small but data cover only a restricted part of state space. Here I condition on the fitted weights and focus on learning the distribution of latent paths.

SDEs extend ordinary differential equations (ODEs) by adding stochastic noise components, typically modeled as Wiener processes. In an It\^o model the drift specifies the local mean rate of change, and the diffusion specifies the local covariance of increments. The mathematical theory builds upon the It\^o integral and It\^o's formula \citep{sarkka2019applied}. Standard Brownian motion illustrates the difference between local moment descriptions and sample-path smoothness. Its increments over a time interval of length \(h\) have variance \(h\), yet almost every path is nowhere differentiable. For that reason, any energy term used below must be defined for an admissible drift or control field, rather than in terms of the ordinary velocity of a realized diffusion path.

The numerical approximation of SDEs requires care because convergence of sample paths and convergence of expectations answer different questions. Euler--Maruyama approximates the drift over a step of length \(h\) by a term of order \(h\) and its Gaussian innovation by a term of order \(\sqrt h\). Under the usual global Lipschitz and growth conditions it has strong order one half. Higher-order schemes, such as Milstein's, use additional terms that depend on the diffusion \citep{kloeden1992numerical}. Throughout the gradient analysis I fix the grid and differentiate the actual discrete map. Refinement changes the objective as well as the random gradient. The empirical grid comparison therefore evaluates each correction relative to its own uncorrected discrete gradient.

Incorporating SDEs into machine learning has gained traction with the advent of neural SDEs, where the drift and diffusion terms are parameterized by neural networks. Neural ODEs \citep{Chen2018} and latent ODEs \citep{rubanova2019latent} established ways to learn continuously evolving hidden states for irregularly sampled series. A latent ODE can propagate uncertainty in its initial state, whereas latent SDE dynamics introduce innovations throughout the path. Neural SDEs have also been trained as continuous-time generative models using adversarial objectives \citep{kidger2021neural}. The distinction in the present paper is between the generative diffusion, which specifies a prior path law, and the variational smoother, which changes that law in light of observations.

Variational inference over diffusion paths predates latent neural SDEs \citep{archambeau2007variational}. \citet{li2020scalable} combine shared-diffusion variational inference with stochastic adjoint differentiation. Observation guides also appear in guided diffusion bridges, where an auxiliary process approximates the effect of endpoint conditioning \citep{schauer2017guided}. I use a shared-diffusion posterior for the same tractability reason as earlier latent SDE work. The question specific to SLDI is how much evidence is lost if its encoder changes only the initial law. This can be answered exactly even though the unrestricted smoothing distribution may be intractable.

Methods for differentiating SDE solutions address memory use, numerical adjoint accuracy, and the variance of the gradient returned by a solver. \citet{kidger2021efficient} study numerical accuracy and reversibility. \citet{roeder2017sticking} remove a centered score contribution from a reparameterized gradient, and \citet{xu2022infinitely} adapt the idea to SDEs. \citet{miller2017reducing} form control variates using tractable approximations to reparameterization gradients. Adjoint matching also uses sensitivities as regression targets in stochastic control \citep{dom2025adjoint,mei2026bsde}. My objective is the discrete variational gradient for an irregular observation sequence. The realized adjoint computes that gradient, and the learned component predicts a correction whose coefficient is fixed before the current solver innovation.

The closest mathematical antecedent is \citet{belomestny2018variance}, who use an orthogonal expansion in step innovations to reduce the variance of discretized diffusion functionals. For an ELBO gradient, the conditional quantity to approximate depends on both the latent state and the derivative of that state with respect to the selected parameter directions. I show how to use mixed adjoints to estimate the first two orders of that expansion. This also gives a way to attribute variance changes to individual steps. The experiments use that attribution to compare two equally sized sets of corrections, at the first and last steps of each observation interval.

\section*{Methodology}

\subsection*{Latent dynamics and variational inference}

I begin by modeling latent variables $\{z_t\}_{0\leq t\leq T}$ evolving in continuous time as the solution to an It\^o stochastic differential equation (SDE). Specifically, let $z_t \in \mathbb{R}^d$ denote the latent state at time $t$. The evolution of $z_t$ is governed by:

\[
dz_t = \mu_\theta(z_t, t)\,dt + \Sigma_\theta(z_t, t)\,dW_t,
\]
where $\mu_\theta : \mathbb{R}^d \times \mathbb{R} \rightarrow \mathbb{R}^d$ is a drift function, $\Sigma_\theta : \mathbb{R}^d \times \mathbb{R} \rightarrow \mathbb{R}^{d \times m}$ is a diffusion coefficient, and $W_t \in \mathbb{R}^m$ is an $m$-dimensional Brownian motion. The drift and diffusion terms are parameterized by neural networks with parameters $\theta$ to be optimized through variational inference.

This stochastic process induces a distribution over sample paths in the latent space, denoted by the probability measure $\mathbb{P}_\theta$ on the path space $\mathcal{C}([0,T], \mathbb{R}^d)$. With the usual well-posedness conditions, the coefficients specify a Markov process with time-dependent transition uncertainty. For fixed $\theta$, the marginal distribution $p_\theta(z_t)$ evolves according to the Fokker--Planck equation \citep{sarkka2019applied}, which governs the time evolution of the probability density function $\rho(z, t)$ of $z_t$:

\[
\frac{\partial \rho}{\partial t} = -\nabla_z \cdot (\mu_\theta(z, t) \rho) + \frac{1}{2} \sum_{i,j=1}^d \frac{\partial^2}{\partial z_i \partial z_j} \left( D_{ij}(z, t) \rho \right),
\]
where $D(z, t) = \Sigma_\theta(z, t) \Sigma_\theta(z, t)^\top$ is the diffusion matrix. This partial differential equation highlights how both deterministic drift and stochastic diffusion shape the evolution of uncertainty in latent space. The density equation describes the marginal law, while the method below works directly with draws of the latent path.

To numerically simulate sample paths of the latent process, I adopt the Euler--Maruyama discretization \citep{kloeden1992numerical}. Letting $\Delta t$ be a small time step, the discretized update reads:

\[
z_{t+\Delta t} = z_t + \mu_\theta(z_t, t) \Delta t + \Sigma_\theta(z_t, t) \sqrt{\Delta t} \cdot \epsilon_t, \quad \epsilon_t \sim \mathcal{N}(0, I_m),
\]
where $\epsilon_t$ is a standard Gaussian noise vector in $\mathbb{R}^m$. The same discretization is used for generation and recognition, with the respective drift in each case. Euler--Maruyama is practical when exact simulation is infeasible. The time grid includes every emission time.

For integrable moments and under the usual conditions for It\^o's formula, the first and second moments satisfy

\begin{align*}
\frac{d}{dt} \mathbb{E}[z_t] &= \mathbb{E}[\mu_\theta(z_t, t)], \\
\frac{d}{dt} \text{Cov}(z_t) &= \mathbb{E}[\Sigma_\theta(z_t, t) \Sigma_\theta(z_t, t)^\top] + \Cov(z_t,\mu_\theta(z_t,t))+\Cov(\mu_\theta(z_t,t),z_t),
\end{align*}
where \(\Cov(A,B)=\E[(A-\E A)(B-\E B)^\top]\). The covariance formula follows by applying It\^o's product rule to \(z_tz_t^\top\) and differentiating \(\E[z_t]\E[z_t]^\top\). The two cross covariances can have either sign. For instance, the drift of an Ornstein--Uhlenbeck process contracts the covariance while its diffusion increases it. The quadratic variation, \(\int_0^tD(z_s,s)\,ds\), instead describes the accumulation of infinitesimal path variability. This distinction is important when interpreting estimated process noise.

The model uses a prior \(p_0\), typically a standard Gaussian, and the generative path law will be denoted \(P_\theta\). Given observation times \(0<t_1<\dots<t_M\leq T\), I write \(x=\{(t_i,x_i)\}_{i=1}^M\) for their values and timestamps. The likelihood is \(\ell_x(z)=\prod_i p_\psi(x_i\mid z_{t_i})\), conditional on the given schedule. If the observation times are informative in their own right, a generative model would also be needed for those times. Treating the schedule as fixed here isolates the smoothing problem given the observations.

Given the latent trajectory, I define the observation model through a conditional likelihood. In its simplest form, I use a multivariate Gaussian with a neural mean \(f_\psi\) and a fixed isotropic variance:
\[
 p_\psi(x_i\mid z_{t_i})=\N(x_i; f_\psi(z_{t_i}),\sigma^2 I).
\]
A feedforward decoder suffices for scalar or tabular measurements. A convolutional decoder can be appropriate for image observations. When measurement variability changes with the state, the network can also output a positive diagonal scale, giving
\[
 p_\psi(x_i\mid z_{t_i})
 =\N\bigl(x_i;\mu_\psi(z_{t_i}),\diag(\sigma_\psi^2(z_{t_i}))\bigr).
\]
For categorical or count data the emission distribution can be replaced by a categorical or Poisson likelihood. Separating the observation likelihood from the stochastic transition is useful because both contribute to conditional predictive variability. The observed sequence alone may give limited information about their relative contributions. The experiments hold the observation noise fixed so that the effect of the estimator can be isolated.

An encoder in the original SLDI formulation first processes the full observation sequence and returns a Gaussian \(q_{\phi,0}(\cdot\mid x)\) over the latent initial state. A bidirectional recurrent network can handle this task by taking the time stamps or gaps as inputs. The generative dynamics propagate the sampled initial condition. This gives the initial-only law \(Q(dz)=q_0(dz_0)P_\theta(dz\mid z_0)\), whose conditional kernel matches that of the prior. Its path KL is therefore the initial KL. The best achievable bound for such an encoder can be found without assuming any neural parameterization of \(q_0\).

Let \(m_x(z_0)=\E_{P_\theta}[\log \ell_x(Z)\mid Z_0=z_0]\). Suppose \(\ell_x>0\) almost surely, \(\log\ell_x\in L^1(P_\theta)\) and \(p(x)=\E_{P_\theta}\ell_x(Z)<\infty\). Define \(A_x=\E_{p_0}e^{m_x(Z_0)}\). If the tilted law below has finite initial KL and finite expected absolute log likelihood, the maximal initial-only ELBO is \(\log A_x\). It is attained at
\[
 q_0^\star(dz_0)=A_x^{-1}e^{m_x(z_0)}p_0(dz_0).
\]
The gap \(\log p(x)-\log A_x\) is nonnegative. It is zero exactly when \(\ell_x(Z)\) is almost surely constant under \(P_\theta(\cdot\mid z_0)\) for \(p_0\)-almost every \(z_0\).

The ELBO for an admissible \(q_0\) is \(\E_{q_0}m_x-\KL(q_0\|p_0)=\log A_x-\KL(q_0\|q_0^\star)\). Conditional Jensen gives \(e^{m_x(z_0)}\le \E_{P_\theta}[\ell_x\mid Z_0=z_0]\), so \(0<A_x\le p(x)\). Strict convexity of the exponential gives the stated conditional equality criterion. The integrated difference of the two conditional expressions can vanish only when their pointwise difference vanishes almost surely.

The equality condition has a useful interpretation. The exact posterior initial law would reweight \(p_0\) by the arithmetic conditional likelihood \(\E_{P_\theta}[\ell_x\mid z_0]\). The optimal initial-only variational law instead uses the exponential of the average log likelihood. Their gap measures how much the fixed conditional path kernel loses by ignoring observation information about the post-initial innovations. It persists even if the family of encoders can represent every initial distribution. The degeneracy condition covers cases where the likelihood depends only on the initial state. With process noise and an informative measurement later in time, that condition is generally violated. The scalar Gaussian example below quantifies the gap for a case in which it can be computed exactly.

For smoothing, an encoder reads the full observed sequence with timestamps. It outputs an initial distribution \(q_{\phi,0}(\cdot\mid x)\) and a deterministic context \(c_x(t)\). This context may be produced by a bidirectional sequence encoder. I let a recognition drift \(u_\phi\) depend on this context to update the latent dynamics along the path.
\[
 Z_0\sim q_{\phi,0},\qquad
 dZ_t=[\mu_\theta(Z_t,t)+\Sigma_\theta(Z_t,t)u_\phi(Z_t,t,c_x(t))]\,dt
       +\Sigma_\theta(Z_t,t)\,dW_t^Q.
\]
I assume strong well-posedness, equivalent initial distributions with finite initial KL, a square invertible diffusion, and a valid Girsanov change of measure with finite expected control energy under \(Q\) (for example, bounded \(u_\phi\) on a finite horizon). Invertibility makes the driving innovations recoverable from the latent path. Girsanov's formula \citep{girsanov1960transforming} gives
\[
 \KL(Q\|P_\theta)=\KL(q_{\phi,0}\|p_0)+
   \half\,\E_Q\int_0^T \|u_\phi(Z_t,t,c_x(t))\|^2\,dt .
\]
The log density ratio under \(Q\) is the initial log density ratio plus the control energy and \(\int u_\phi^\top dW^Q\). The stochastic integral has zero mean. This entropy uses the drift correction in Brownian coordinates. It is finite under the stated assumptions even though diffusion trajectories generally have no time derivative. Setting \(u_\phi=0\) recovers the initial-only model. The smoothing context can depend on future observations relative to physical time because the sequence \(x\) is fixed before simulating \(W^Q\). For forecasting, the encoder uses only observations available at the forecast origin, and future predictions propagate from the resulting filtering distribution under the generative dynamics.

The exact smoother supplies a reference for what the recognition guide must learn. For a time strictly between observations, let \(h_x(t,z)\) be the expected product of the remaining likelihood factors under the prior conditioned on \(z_t=z\). If this function is strictly positive and sufficiently smooth on each open observation interval, the transition kernel of the posterior up to the next observation is obtained by reweighting the prior transitions by the future likelihood. Its infinitesimal generator can be found by applying the prior generator to the product \(h_x f\) and subtracting \(f\) times the generator applied to \(h_x\), then dividing by \(h_x\). The diffusion remains \(D\), while the drift acquires the term \(D\nabla_z\log h_x\). The corresponding control in Brownian coordinates is
\[
 u^*(t,z)=\Sigma_\theta(z,t)^\top\nabla_z\log h_x(t,z).
\]
The remaining-likelihood function satisfies a backward equation within each open interval. At an observation time, its left limit includes that observation's likelihood factor and its right limit includes the later factors. The construction uses noisy likelihood factors, so the latent path stays continuous through observation times. The resulting drift depends on both the time and the available observation context. Its expression can guide the design of the recognition model even when the remaining-likelihood function is intractable. Approximating this drift, learning it from data, and choosing the numerical grid remain distinct problems.

The SLDI architecture allows these components to be varied independently. A recurrent encoder can produce an initial mean and scale and a time-indexed sequence of recognition features. The drift network takes the current latent state and the requested simulation time, with weights shared across time. A diagonal diffusion can be formed by mapping network outputs to strictly positive scales, for example by adding a positive floor to a softplus. A bounded final activation in the guide is one simple way to control its energy over a finite horizon. These are parameterization choices. Strong well-posedness also depends on the resulting functions of the state, and a positive diagonal by itself does not impose a growth bound.

In an implementation the emission contribution is evaluated only when the integration grid reaches the relevant observation. A context based on the next measurement can be held constant between successive observations, as in the small nonlinear experiment. A richer bidirectional encoder can instead use information from the whole later subsequence. If one refines the solver grid, the observation times and their likelihood factors should stay fixed. This avoids an artificial change in the number of emission terms when only the integration accuracy has changed.

The control energy is the pathwise regularizer in the ELBO. It penalizes the change of drift in the coordinates in which the noise is standard. The quadratic action based on \(\dot z_t\) would instead attempt to penalize the ordinary path velocity. To see the numerical effect of such a substitution, take a scalar zero-drift diffusion with constant \(\sigma\). The expected sum of squared difference quotients multiplied by their step lengths is \(N\sigma^2\) on any partition of \(N\) steps. That quantity diverges with refinement even though the generative and recognition laws can coincide. By contrast, the control energy vanishes in that case. This regularization has its coefficient fixed by relative entropy. Adjusting the coefficient would define a different objective.

Fix a grid \(s_0=0<\dots<s_N=T\) containing all observation times, set \(\Delta_n=s_{n+1}-s_n\), and write \(w\) for all trainable parameters. The recognition step is
\[
 Z_{n+1}=F_n(Z_n,\xi_n;w)
 =Z_n+\Delta_n[\mu_\theta+\Sigma_\theta u_\phi]_n
   +\sqrt{\Delta_n}\,\Sigma_{\theta,n}\xi_n,\qquad
 \xi_n\stackrel{\mathrm{iid}}{\sim}\N(0,I).
\]
The prior Euler step uses \(u_\phi=0\) and has the same conditional Gaussian covariance. The chain rule for relative entropy therefore yields an \emph{exact} ELBO for the Euler model, with the conditional transition KL \(\Delta_n\|u_\phi\|^2/2\). Let \(\ell_n(z;w)\) be the log likelihood of emissions at \(s_n\), zero when there are none, and let \(k_n(z;w)=-\Delta_n\|u_\phi(z,s_n,c_x(s_n))\|^2/2\). Then
\[
 J_\Delta(w)=\E\!\left[L_\Delta\right],\qquad
 L_\Delta = -\KL(q_{\phi,0}\|p_0)+
             \sum_{n=0}^{N-1}\bigl[k_n(Z_n;w)+\ell_{n+1}(Z_{n+1};w)\bigr].
\]
This form uses a tractable initial KL, for example with Gaussian initial laws. An initial reparameterization \(Z_0=H(\zeta;w,x)\) uses an independent base draw \(\zeta\). The gradient and variance results below concern \(J_\Delta\). The difference between this objective and the continuous-time objective involves solver accuracy and is outside the variance correction.

The conditional Gaussian calculation also explains the role of using a shared diffusion in the Euler model. At a common state the transition means differ by \(\Delta_n\Sigma_{\theta,n}u_{\phi,n}\), while the covariance is \(\Delta_n\Sigma_{\theta,n}\Sigma_{\theta,n}^\top\). The Gaussian mean-shift KL is one half of the squared Mahalanobis norm of the mean difference, which simplifies to \(\Delta_n\|u_{\phi,n}\|^2/2\). Thus the path KL used in the finite-grid objective arises from the transitions actually used for sampling. Both evidence and the ELBO depend on the grid. In particular, the same diffusion coefficient in continuous time allows a change in instantaneous drift under posterior conditioning, while conditioning a finite Gaussian transition can change its mean and its covariance. The numerical Gaussian example below computes the resulting residual evidence gap separately from the restriction to the initial state.

\subsection*{Adjoint-based backpropagation and variance reduction}

Training stochastic processes with continuous-time dynamics poses a computational challenge. The original SLDI architecture assigned an auxiliary network to model the adjoint alongside the forward process. I retain an auxiliary field and define its target through the derivative of the exact discrete calculation. The ordinary adjoint carries the sensitivity of the ELBO through the latent steps. The learned field instead estimates which part of that gradient varies predictably with each innovation. The two calculations answer distinct questions. In particular, memory savings from recomputing or reversing a solver step are independent of the variance gains from a learned correction.

Assume \(L_\Delta\) is differentiable in \(w\), differentiation can pass through the expectation in the discrete ELBO, and \(g=\nabla_w L_\Delta\) is square integrable. The ordinary discrete gradient is computed by reverse-mode differentiation of the Euler recurrence. With column adjoints,
\begin{align*}
 a_N &= \nabla_z\ell_N(Z_N;w), \\
 a_n &= \nabla_z(\ell_n+k_n)(Z_n;w)+(D_zF_n)^\top a_{n+1},
 \qquad  n=N-1,\ldots,0,\\
 g &= -\nabla_w\KL(q_{\phi,0}\|p_0)+(D_wH)^\top a_0 \\
 &\quad+\sum_{n=0}^{N-1}\bigl\{(D_wF_n)^\top a_{n+1}
   +\partial_w[k_n(Z_n;w)+\ell_{n+1}(Z_{n+1};w)]\bigr\}.
\end{align*}
Here the Jacobians of \(F_n\) are evaluated on the simulated path, partial parameter derivatives hold the latent state fixed, and \(\ell_0=0\). This recurrence is the adjoint of the implemented random map, including the state derivatives of the diffusion coefficients and the impulse contributions of emissions.

Choose an orthonormal \(p\times r\) matrix \(V\) of parameter directions, with \(p=\dim w\), and let \(G=V^\top g\). For example, \(V\) can select a small block of the recognition network. All statements hold conditionally on the fixed data, parameters and directions. The pre-step filtration is \(\F_n=\sigma(\zeta,\xi_0,\ldots,\xi_{n-1})\). Let \(\mathcal H\) be a finite set of nonconstant, orthonormal Hermite polynomials for the standard Gaussian law, indexed by multi-indices \(\alpha\). Define the correction
\[
 C=\sum_{n=0}^{N-1}\sum_{\alpha\in\mathcal H}
      \widehat A_{n,\alpha}(U_n)h_\alpha(\xi_n),
 \qquad U_n\text{ is }\F_n\text{-measurable}.
\]
The coefficients are fitted on independent pilot paths or fixed by training history before the current noise is drawn. Treat \(C\) as a value to subtract after computing \(g\). Thus \(g-VC\) corrects the projected directions and retains the exact sample derivative in the orthogonal complement. In an autodiff implementation the fitted correction is detached from the update.

The correction is conditionally unbiased. To see this, fix an independently fitted set of square-integrable coefficients, and take expectations over a fresh path. Then
\(\E[g-VC]=\nabla_w J_\Delta(w)\).
Let
\[
 B_{n,\alpha}=\E[G h_\alpha(\xi_n)\mid\F_n],
 \qquad A_{n,\alpha}(U_n)=\E[B_{n,\alpha}\mid U_n].
\]
Writing \(\mathcal H^c\) for the omitted nonconstant Hermite indices, the adjusted projected gradient satisfies
\begin{align*}
\tr\Cov(G-C) &=
 \tr\Cov(\E[G\mid\F_0])+
 \sum_{n,\alpha\in\mathcal H^c}\E\|B_{n,\alpha}\|^2
 \\
 &\quad+\sum_{n,\alpha\in\mathcal H}\E\|B_{n,\alpha}-A_{n,\alpha}(U_n)\|^2
       +\sum_{n,\alpha\in\mathcal H}\E\|A_{n,\alpha}(U_n)-\widehat A_{n,\alpha}(U_n)\|^2.
\end{align*}
The optimal coefficients given \(U_n\) are \(A_{n,\alpha}\). The change in full-gradient covariance trace equals the change in the projected trace.

Each correction has conditional mean zero given \(\F_n\) and the fit, by independence of the fresh innovation. Let \(M_n=\E[G\mid\F_n]\). Since \(G\) is \(\F_N\)-measurable, it equals \(M_0+\sum_n(M_{n+1}-M_n)\). Conditional on \(\F_n\), expand each difference in the complete Hermite basis of the independent \(\xi_n\). Its coefficient is \(B_{n,\alpha}\) and its constant term is zero. Terms from different steps are orthogonal because they are martingale differences, and terms at the same step are orthogonal by the conditional Hermite inner products. The remaining within-index squared error splits by conditional expectation on \(U_n\), giving the residual-variance formula. Finally, the squared Euclidean norm of a centered gradient decomposes into its projections onto the span of \(V\) and its orthogonal complement.

The residual-variance formula leaves three errors for the fitted correction: omitted Hermite orders, information discarded by \(U_n\), and coefficient estimation error. A fit can increase variance if its estimation error exceeds the reduction achieved by the population coefficients. Any validation-based decision to include or scale a correction must also be made independently of the current path. The separation matters even for a single step: with \(G(\xi)=\xi^2\), using \(\widehat A(\xi)=\xi\) in the correction \(\widehat A(\xi)\xi\) shifts the gradient mean by one. A coefficient network evaluated on a post-step state or a realized backward adjoint can produce this dependence.

Let \(S_n=(D_w Z_n)V\) be the tangent of the state in the corrected directions, and define \(X_n=(Z_n,S_n)\). It is computed before step \(n\) by
\[
 S_0=(D_w H)V,\qquad
 S_{n+1}=D_zF_n\,S_n+D_wF_n\,V .
\]
For a fixed input sequence and grid, \(X_n\) is a Markov state. The contribution to \(G\) of all costs accrued before drawing \(\xi_n\) is \(\F_n\)-measurable. Their product with any nonconstant \(h_\alpha(\xi_n)\) has conditional expectation zero. All other gradient contributions are deterministic functions of \(X_n\) and the remaining innovations, using the tangent recurrence and the directional derivatives of the step costs. Therefore \(B_{n,\alpha}\) depends on the past only through \(X_n\). Using \(U_n=X_n\) eliminates the feature error in the residual-variance formula at the population optimum. A coefficient network shared across sequences and steps would also take the observation context and grid information needed to specify the remaining recurrence as inputs.

The direct regression target is \(G h_\alpha(\xi_n)\). At low Hermite orders we can instead use Gaussian integration by parts. Let \(D_i G=\partial G/\partial(\xi_n)_i\) with all other base draws fixed. Under conditional Gaussian integration-by-parts regularity (integrability of the displayed targets and vanishing boundary terms), the following targets have the same conditional mean:
\[
\begin{array}{c|c}
 h_\alpha(\xi_n) & \text{mixed-derivative regression target}\\ \hline
 (\xi_n)_i & D_iG\\[2pt]
 ((\xi_n)_i^2-1)/\sqrt{2} & (\xi_n)_i D_iG/\sqrt{2}\\[2pt]
 (\xi_n)_i(\xi_n)_j,\quad i<j &
 \bigl[(\xi_n)_j D_iG+(\xi_n)_i D_jG\bigr]/2 .
\end{array}
\]
For the diagonal term, apply the first-order identity to \(G(\xi_n)_i\), then subtract the expectation of \(G\). For the cross term, apply it in either coordinate and average. The derivatives are taken before projecting onto \(U_n\), so the same target identity holds with state-only features. The target paths can use all future innovations. Only the fitted prediction used in the predictable correction must be predictable.

The derivatives in the mixed-target display can be obtained by a single reverse sweep of the directional-gradient calculation for each scalar output, or by a batched reverse sweep for the chosen block. Specifically write \(X_{n+1}=\mathcal F_n(X_n,\xi_n)\) for the augmented recurrence, and write \(G=\mathrm{const}+\sum_{n=0}^N R_n(X_n)\) using the directional derivatives of \(\ell_n+k_n\), with \(k_N=0\). The constant is the derivative of the analytic initial KL. With \(P_n\) a matrix of \(r\) adjoint columns,
\[
 \begin{aligned}
 P_N&=(D_XR_N)^\top,\qquad
 P_n=(D_XR_n)^\top+(D_X\mathcal F_n)^\top P_{n+1},\\
 (D_{\xi_n}G)^\top&=(D_{\xi_n}\mathcal F_n)^\top P_{n+1}.
 \end{aligned}
\]
In the last line the step cost \(R_n\) has no explicit dependence on \(\xi_n\). This sweep uses mixed second derivatives of the original computation and supplies both first- and second-order targets without an innovation Hessian of \(G\). If pilot computing cost is the limiting factor, direct moment targets remain an alternative. At deployment, the update requires the ordinary gradient, the predictable features and the fitted correction. It need not compute mixed derivatives on the fresh path.

At a fixed parameter iterate, simulate independent pilot paths and evaluate the tangent, innovations and mixed targets. Fit the coefficient functions by squared loss. On a fresh batch, evaluate these functions immediately before each step and accumulate the predictable correction using the subsequent innovation. Compute the discrete adjoint and average the corrected gradients for the model update. A fit from a preceding iteration preserves conditional unbiasedness if it is fixed before drawing the new batch, although stale coefficients can increase the variance. Pilot and update draws must remain independent conditional on the training history.

A practical issue is that the regression can be easier in some parameter directions than in others. The tangent state has one column per corrected direction. For a nonlinear drift network it can be more practical to choose a block such as the output weights of the recognition field, using the ordinary sample gradient in the remaining coordinates. The coefficients are trained with a supervised squared error. Their approximation budget includes the numerical conditioning of the chosen features, the order of the polynomial basis, and the number of independent pilot trajectories. A fixed small model is useful for the numerical study because it allows the state-only and tangent-augmented regressions to use the same polynomial family and the same training paths.

The martingale construction also gives an experiment with a time-local interpretation. Denote the fitted correction at a single solver step by \(C_n\) and let \(I\) be any deterministic subset of steps for the given observation sequence. When the coefficients are fixed and the features are pre-step, the correction restricted to \(I\) remains centered. Orthogonality between distinct steps gives the expected reduction in covariance trace:
\[
 \tr\Cov(G)-\tr\Cov\left(G-\sum_{n\in I}C_n\right)
   =\sum_{n\in I}\left(2\,\E[(G-\E G)^\top C_n]-\E\|C_n\|^2\right).
\]
This expression lets me compare subsets with the same number of corrections. At the population projection the summand is nonnegative and equals the energy of the projected component at that step. For a fitted correction the summand can also be negative, for example when the fitted term has substantial variation and little correlation with the gradient. The formula concerns different time steps, not different observation sequences. Aggregating the reductions across sequences still depends on how those sequences are weighted. In the experiments, I normalize the reductions within each context and seed by the uncorrected gradient trace.

A second way to read the formula is as a diagnostic for grid refinement. As an observation interval is split into more steps, a correction applied to just one of those steps covers a smaller fraction of the driving Gaussian noise. The correction over the entire interval can nevertheless retain its variance benefit if the individual steps remain predictable from the chosen features. Between those extremes there is an empirical question about whether some steps are disproportionately important. I use the first and last step in each interval as paired interventions. This comparison retains the rest of the simulated path and its exact gradient, so a difference between subsets measures where the correction acts rather than a change in the numerical solution.
\subsection*{Numerical experiments}

I use two synthetic problems to evaluate the smoothing restriction and the predictable adjoint correction. The linear problem fixes the exact path law, giving independent checks of the best initial-only bound and of a gradient whose innovation variance can be removed algebraically. The nonlinear problem uses irregular observations, a state-dependent diffusion, and a nonzero recognition guide. It tests the fitted rather than the oracle correction. I first study the target and feature choices at different pilot budgets, then vary the integration grid and test how much of the variance reduction is concentrated near emissions. The tests condition on the chosen observations and compare estimators of the same grid-specific gradient.

For the Euler discretization of the scalar Ornstein--Uhlenbeck process with drift \(-a Z_t\) and diffusion \(\sigma\), let \(\rho=\prod_n (1-a\Delta_n)\) and \(\beta_n=\sqrt{\Delta_n}\prod_{j>n}(1-a\Delta_j)\). Then \(Z_N=\rho Z_0+\sigma W\), with \(W=\sum_n\beta_n\xi_n\) and \(B=\sum_n\beta_n^2\). There is one noisy observation \(x\mid Z_N\sim\N(Z_N,\tau^2)\).

With \(Z_0\sim\N(m_0,v_0)\), let \(\varphi(x;m,v)\) denote the univariate Gaussian density. The exact evidence and the restricted bound of the initial-only calculation are
\begin{align*}
 \log p(x)&=\log\varphi(x;\rho m_0,\rho^2v_0+\tau^2+\sigma^2 B),\\
 \log A_x&=\log\varphi(x;\rho m_0,\rho^2v_0+\tau^2)
              -\frac{\sigma^2 B}{2\tau^2}.
\end{align*}
The second display follows by averaging the log likelihood at fixed \(Z_0\) before exponentiating. For \(a=0.65\), \(\sigma=0.75\), \(\tau=0.35\), \(x=0.70\), \(m_0=-0.20\), \(v_0=0.60\), and steps \((0.13,0.29,0.11,0.37,0.41,0.24,0.45)\), the gap is \(2.578\) nats.

The same model makes it possible to isolate the part of the gap that remains after allowing an arbitrary observation-dependent mean shift at each Euler step. Write \(a_n=1-a\Delta_n\) and \(Q_n=\sigma^2\Delta_n\), so the transition from \(Z_n\) has mean \(a_n Z_n\) and variance \(Q_n\). Let \(r_{n+1}=\prod_{j>n}a_j\) and \(R_{n+1}=\tau^2+\sigma^2\sum_{j>n}\beta_j^2\). Given \(Z_{n+1}\), the terminal measurement has conditional mean \(r_{n+1}Z_{n+1}\) and variance \(R_{n+1}\). Completing the square in the product of the Gaussian transition and this remaining likelihood gives the posterior transition with variance \(V_n\) and mean \(m_n(z,x)\):
\[
\begin{aligned}
 V_n^{-1} &= Q_n^{-1}+r_{n+1}^2/R_{n+1},\\
 m_n(z,x) &= V_n\bigl(a_n z/Q_n+r_{n+1}x/R_{n+1}\bigr).
\end{aligned}
\]
An unrestricted drift correction in the Euler family can match this conditional mean. Its variance is still \(Q_n\). The reverse KL between two normal laws at a given state is minimized over the recognition mean by matching the posterior mean. The resulting penalty is independent of the state and observation value. Because each such penalty is attainable simultaneously and the initial law can be chosen to match the exact posterior initial law, the chain rule gives the best attainable gap in the drift-only Euler family:
\[
 \frac12\sum_n\left(\gamma_n-\log(1+\gamma_n)\right),
 \qquad \gamma_n=\frac{Q_n r_{n+1}^2}{R_{n+1}} .
\]
For the same grid and noise parameters, this gap is \(0.489\) nats. I checked it independently using the KL of the two joint Gaussian path vectors. Thus the unrestricted mean shifts remove most of the initial-only gap in this example, and the remaining gap is explained by the Euler transition covariance constraint. The two numbers compare variational families at their exact optima on one grid. The nonlinear guide in the next experiment is a finite parameterization whose actual evidence gap is unknown.

The covariance penalty has an additional useful property. For \(\gamma\geq 0\), \(\gamma-\log(1+\gamma)=\int_0^\gamma s/(1+s)\,ds\leq\gamma^2/2\). With positive observation variance and a fixed time horizon, the factors \(r_{n+1}\) stay bounded under refinement and \(R_{n+1}\geq\tau^2\). Each \(\gamma_n\) is then \(O(\Delta_n)\), and the sum of covariance penalties is \(O(\max_n\Delta_n)\). This finite-grid calculation illustrates how shared instantaneous diffusion is consistent with posterior conditioning even though shared finite-step covariance is restrictive. The derivation relies on a noisy terminal observation and on the linear transition. Its rate does not directly apply to state-dependent diffusion or exact endpoint constraints.

For the gradient check I take the known initial state \(z_0=m_0\) and \(u=0\). Let \(p_n=\sum_{j<n}\beta_j\xi_j\). The sample derivative with respect to \(\sigma\) is
\[
 G=\frac{(x-\rho z_0)W-\sigma W^2}{\tau^2}.
\]
The conditional first- and second-order coefficients for step \(n\) are
\[
 B_{n,1}=\frac{\beta_n(x-\rho z_0-2\sigma p_n)}{\tau^2},
 \qquad
 B_{n,2}=-\frac{\sqrt{2}\,\sigma\beta_n^2}{\tau^2}.
\]
They follow by expanding \(W=p_n+\beta_n\xi_n+\sum_{j>n}\beta_j\xi_j\). All higher coefficients vanish. The raw and first-order residual variances are
\[
 \Var(G)=\frac{(x-\rho z_0)^2 B+2\sigma^2 B^2}{\tau^4},
 \qquad
 \Var(G-C_1)=\frac{2\sigma^2\sum_n\beta_n^4}{\tau^4}.
\]
These are \(83.28\) and \(18.32\) for the chosen grid. On \(200{,}000\) simulated paths the corresponding variances were \(82.87\) and \(18.58\). With both orders the adjusted gradient was within \(3.2\times 10^{-14}\) of the exact conditional mean \(-\sigma B/\tau^2\) on every simulated path.

The Gaussian oracle also shows how an individual step can matter despite independent raw innovations. For \(d=x-\rho z_0\) and \(B_{<n}=\sum_{j<n}\beta_j^2\), the two-order projected energy of step \(n\) is
\[
 \E[C_n^2]=\frac{\beta_n^2(d^2+4\sigma^2 B_{<n})+2\sigma^2\beta_n^4}{\tau^4}.
\]
The first Hermite coefficient uses the running prefix \(p_n\), so it can remove cross-products involving previous innovations. The second coefficient cancels the remaining centered squares. A large first coefficient can reflect both the propagation weight \(\beta_n\) and uncertainty accumulated earlier. This prediction is specific to the linear terminal-observation problem. In nonlinear smoothing the relative contributions may vary with the observation schedule as well as the dynamics and parameter block.

The scalar latent model has \(Z_0=-0.2\), \(b(z)=-a z+b_1\tanh(z)\), and \(\Sigma(z)=0.55+0.19\tanh(z)\). Emissions are Gaussian with mean \(Z_{t_i}\) and standard deviation \(0.30\). The recognition guide is \(u_\phi(z,t,c_x)=c\tanh(y_{\rm next}-z)\), where \(y_{\rm next}\) is the next scheduled observation value from the fixed sequence. The guide is held to that observation on each grid interval leading up to its emission. Four contexts were simulated independently with the prior using \(a=0.8\), \(b_1=0.55\) and 16 Euler substeps per observation interval. Each context has four irregular times drawn uniformly over \([0.20,1.80]\) and a final time of \(2.0\). For the estimator comparison I use four substeps per interval and evaluate the gradient in all three coordinates \((a,b_1,c)\) at \((0.8,0.55,0.9)\). The fixed prior and recognition initial states coincide, so their initial KL is zero.

At each step in each context, we fit a quadratic ridge regression on standardized nonconstant polynomial features of \((Z_n,S_n)\). The penalty coefficient is \(0.01\) when squared errors are averaged and the intercept is unpenalized. I compare direct moment targets through order two, the mixed targets through orders one and two, and a second-order mixed-target control with only \(Z_n\) as input. I also compare antithetic sampling by pairing a path at \(\xi\) with one at \(-\xi\); its pair-mean variance is multiplied by two to account for the two path evaluations. The other ratios count only fresh evaluation paths. The pilot sizes are 128 and 1,024, using nested pilot draws, and the evaluation size is 8,192 independent paths per context and seed. The base estimator and each correction use the same evaluation paths. Both the parameter gradient and the mixed derivatives were checked against centered finite differences. I aggregate eight seeds in each of four contexts.

With 128 pilot paths, the first-order mixed-target correction retained a median \(8.1\%\) of the raw covariance trace, compared with \(11.2\%\) for the two-order fit and \(12.6\%\) for the two-order state-only fit. With 1,024 pilot paths, the corresponding medians were \(4.7\%\), \(1.5\%\), and \(8.1\%\). The interquartile range for the two-order tangent fit was \(9.2\%\) to \(13.9\%\) at the smaller pilot size and \(1.3\%\) to \(1.6\%\) at the larger one. Direct moment fits with the same Hermite basis retained median traces of \(598\%\) and \(106\%\) at these pilot sizes. The antithetic comparison, expressed per path evaluation, retained \(137\%\). This ordering is consistent with the variance decomposition. At the population optimum an additional Hermite order enlarges the projection space. With finite data each additional coefficient also has to be estimated, and the larger pilot sample is what makes the second order effective in this experiment.

I next used a separate ensemble of pilot and evaluation draws for the grid comparison. The observations and fitted model parameters are held fixed. Each observation interval is discretized into two, four, or eight Euler steps. To match the random inputs, I draw the normalized Gaussian increments on the finest grid, then sum the corresponding increments and normalize them to construct the coarser innovations. Each marginal grid still has the intended innovation law. The procedure couples the raw estimators across grids without sharing paths between coefficient estimation and evaluation. At each grid resolution I fit a fresh second-order correction using 1,024 pilot paths and evaluate it on 4,096 held-out paths for each of eight seeds and four contexts.

When all steps were corrected, the tangent fit retained median traces of \(1.29\%\), \(1.47\%\), and \(1.46\%\) at the two-, four-, and eight-step resolutions. The state-only fit retained \(8.12\%\), \(8.20\%\), and \(8.12\%\), respectively. These percentages use each grid's own baseline, so their stability measures the relative variance reduction and cannot establish that the raw variances or the gradients themselves are constant across resolutions. Using matched Brownian paths addresses variation due to changing the random simulation across grids, while the independent pilot sample preserves centering on the evaluation draw.

To localize the effect, I freeze each full fit and apply only its first-step correction in each observation interval, then repeat with only its last-step correction. The number of corrected steps and the held-out paths are identical. For the tangent fit, the last-step subset retained median traces of \(47.3\%\), \(73.5\%\), and \(86.1\%\) of the grid baseline. The first-step subset retained \(54.9\%\), \(79.9\%\), and \(90.1\%\). The last-step subset had lower variance in 25, 26, and 27 of the 32 paired repetitions, respectively. The median paired advantages were \(7.8\), \(5.5\), and \(3.2\) percentage points. Thus the last-step terms consistently helped more in this comparison, while the advantage decreased as each step covered less of the interval. The near-constant full correction ratios and the weakening effect of any one step are jointly visible only when the grid and the corrected subset are varied independently.

The localized result applies to the tested smoother and observation pattern. The last step includes the immediate emission derivative in the backward calculation. Earlier steps can affect the same emission through their influence on the state and the tangent, and their contributions can differ with the drift, diffusion, and placement of observations. The comparison is consequently evidence of a temporal pattern for this setting rather than a universal ordering of solver steps. Both the trace of the actually corrected gradient and the stepwise covariance contributions are available in the replication code. I use the paired trace ratios as the main reported localization statistic because empirical cross-step covariances can deviate from zero in a finite evaluation sample.

Each pairing has a different purpose. Using identical held-out draws for two corrections on the same grid makes their difference attributable to which coefficient functions are used. Coupling fine and coarse draws provides comparable input trajectories across grids without treating the corresponding Euler gradients as estimates of a common finite-grid quantity. Retaining several independent fits per context exposes sensitivity to the pilot draw. The resulting spread describes repeats of this experiment on the four simulated contexts, and is not a confidence interval for a population of real time series. A broader empirical study could test whether the localization persists as measurement noise varies or when some intervals carry little information about the latent state.

\section*{Discussion}

The two parts of the revision concern different sources of error. The initial-only bound concerns the best variational approximation under a fixed conditional path kernel. Its gap is present before any Monte Carlo estimate is taken. By allowing the recognition drift to read the observations, the model can use later measurements when sampling the latent path. Its actual posterior accuracy then depends on the chosen guide and on the quality of its optimization. The gradient analysis concerns the Monte Carlo estimate for a given finite-grid objective. It leaves this path family and the ELBO unchanged.

At the core of the architecture lies a bidirectional encoder that summarizes the observed sequence, together with drift and diffusion networks that carry the latent state across time. The decoder maps latent states back into observation space. This division continues to be useful when comparing implementations. In particular, diagonal observation noise contributes to the conditional spread at a measurement time, whereas latent diffusion can change uncertainty over an entire gap. A neural Gaussian decoder and a learned diffusion can trade off in explaining observations, and interpretable decomposition of those contributions requires model assumptions or sufficiently informative data. The numerical examples hold the decoder noise fixed.

The auxiliary adjoint field uses another division of responsibilities. The analytic discrete adjoint carries the likelihood contribution at each observation and the propagation through each drift and diffusion step. The field predicts how the resulting gradient changes with the next standardized solver innovation. Its regressions can be inaccurate while the update is still unbiased, provided independence from the current innovation is preserved. This makes the field useful as an auxiliary component of backpropagation. It also means that the predictive ability of the learned field can be assessed using independent pilot and evaluation data before giving it any role in optimization.

One reason to evaluate the fit is that the moment and mixed-derivative targets agree only in conditional expectation. To see how their fluctuations can differ, suppose for a single scalar innovation \(e\) that a gradient coordinate is \(A+B e\), where \(A\) and \(B\) are fixed by the past. The first-order direct target is \(A e+B e^2\), with conditional variance \(A^2+2B^2\). The mixed target for the same coefficient is simply \(B\), with zero conditional variance. In multistep models, the gradient also depends on the future and usually has nonlinear dependence on \(e\), so the mixed derivative retains randomness of its own. For higher Hermite orders it is even multiplied by a function of the current innovation. The analytic first-order example explains how variation unrelated to the desired coefficient can enter a direct target. The pilot-budget experiment measures whether that mechanism matters in the chosen nonlinear smoother.

The same distinction affects the use of the auxiliary model across different observation sequences. A separate regression per time point can learn an average over the future innovations for one fixed context. A shared model intended for unseen sequences would also require context about future observations, ideally in a representation that records both their values and their timing. A single observation gap can have different contributions to the ELBO depending on the current uncertainty and the subsequent emission. Sharing coefficients across intervals by elapsed physical time alone can therefore discard information even if the tangent is retained. In the matched grid comparison I keep context fixed and re-estimate the coefficients for each grid, which measures within-context predictability without confounding it with context amortization.

The grid experiment highlights the distinction between a local and a whole-path correction. Correcting the step immediately before each emission captured more variance than correcting an equal number of steps at the starts of the intervals. With finer grids both subsets captured less, but the all-step ratio remained similar. This is consistent with contributions carried by the immediately observed state as well as smaller contributions accumulated across adjacent steps. The result is a motivation for measuring the temporal distribution of variance under other observation models. It does not imply that dropping the earlier corrections is an efficient default, since their aggregate contribution becomes large as the number of steps increases.

On the practical side, SLDI is intended for domains where uncertainty is both structured and evolving. Examples include climate modeling, irregularly sampled health records, and financial time series. The observation schedule is especially consequential in these settings. Instruments can become unavailable, tests can be ordered in response to a patient's condition, and market data can be recorded only when trades occur. Treating those times as given is adequate for the benchmarks here, which generate them independently of the latent process. Applications with informative observation times would require an additional likelihood for the schedule as part of the inference problem. A forecasting use would also need to report scores on observations unavailable to the encoder at the forecast origin, while a smoothing use can legitimately condition on the complete recorded sequence.

The scale of a variance reduction is distinct from the strength of an inference model. A high predictive uncertainty could reflect uncertainty about the latent trajectory, measurement variability, or a poor fit to the dynamics. Lower variance in the gradient estimate only makes one part of the training procedure less sensitive to the simulated innovations. To assess an application, the predictive distribution would still have to be evaluated against held-out measurements, at horizons and gap lengths that reflect the intended use. The synthetic comparisons isolate the role of the auxiliary adjoint so this assessment can be made separately.

For a larger neural model, each additional direction expands the tangent state, and the number of second-order features grows quadratically with the number of noise coordinates. A block of parameters or a subset of innovation features offers a way to limit that cost. The current results condition on a fixed data sequence and fixed parameters. They quantify variation across simulated paths, and leave open differences across training batches, improvements in optimization time, and changes in forecast accuracy or calibration. The computational overhead of producing mixed targets and refreshing the fit needs to be included in any comparison of training time. Adaptive time grids that depend on new solver noise pose an additional filtration question, so the guarantees here use fixed grids.

\section*{Summary and Conclusion}

Stochastic Latent Differential Inference embeds a continuous-time diffusion in an encoder--decoder model for irregularly observed sequences. The original initial-state recognition mechanism has an exact optimum that can remain strictly below the evidence even when the encoder is unrestricted. The observation-conditioned drift extends that mechanism while retaining a tractable shared-diffusion ELBO on paths and on the numerical grid. A learned coefficient field trained from mixed adjoints supplies a separate improvement in gradient estimation. It controls the contribution of solver innovations without modifying the objective or replacing the exact gradient calculation.

The scalar Gaussian case checks the extreme in which second-order control removes all innovation variance. The nonlinear case illustrates the finite-data trade-off between estimator order and fit quality. It also shows where an equally sized subset of corrections was most effective in the tested setting. Both comparisons are reproducible from the fixed seeds and observation schedules in the repository. These results lay the groundwork for evaluating the approach in larger models, where the cost of fitting the correction and selecting the corrected directions will matter alongside its variance benefit.

The LaTeX source, synthetic observations, numerical experiments, and selected Lean checks are available in the \href{https://github.com/jameskrice7/Stochastic-Deep-Learning}{SLDI repository}.

\bibliographystyle{plainnat}
\bibliography{references}

\end{document}